\documentclass[letterpaper]{article} 
\usepackage[preprint]{aaai2027}  
\usepackage[hyphens]{url}  
\usepackage{graphicx} 
\usepackage{natbib}  
\usepackage{caption} 
\usepackage{amsmath,amssymb}
\usepackage{booktabs}

\newcommand{\ofsupp}{}

\newcommand{\anonalt}[2]{#2}

\newcommand{\clickableurl}[1]{%
  \pdfstartlink attr{/Border[0 0 0]}
                user{/Subtype/Link/A<</Type/Action/S/URI/URI(#1)>>}%
  \textcolor{blue}{\url{#1}}%
  \pdfendlink
}

\newcommand{\blfootnote}[1]{%
  \begingroup
    \renewcommand{\thefootnote}{}%
    \footnote{#1}%
    \addtocounter{footnote}{-1}%
  \endgroup
}

\makeatletter
\let\vc@origlabel\label
\renewcommand{\label}[1]{%
  \vc@origlabel{#1}\vc@origlabel{M-#1}\vc@origlabel{S-#1}%
}
\makeatother

\title{Scaling an Autoregressive Transformer for Single-Cell Generation}

\author{
    Aleksandr Sharipov,
    Yusif Mukhtarov,
    Igor Molybog\textsuperscript{\rm 1}
}
\affiliations{
    \textsuperscript{\rm 1}HawAII, University of Hawai'i at Manoa\\
    sharipov@hawaii.edu, yusif.mukhtarovv@gmail.com, molybog@hawaii.edu
}

\begin{document}
\maketitle

\begin{abstract}
We study a self-supervised generation task for single-cell gene expression vectors: given a set of vectors from a cell type, we aim to generate additional gene expression vectors of that cell type. For this task we characterize both the biological fidelity of the generated gene expression vectors and the scaling behavior of the pretraining loss. The model is a causal transformer paired with a learned quantized VAE tokenizer, trained with a cross-entropy loss. To evaluate the model, we condition it on held-out gene expression vectors of a cell type and generate vectors of gene expression, comparing the resulting distribution over gene expression vectors to the ground truth distribution of that cell type. We study the scaling properties of the proposed architecture by varying the number of trained parameters and the amount of training data. To our knowledge, we find the first jointly-fit two-exponent scaling law and compute-optimal frontier for a single-cell foundation model. Finally, we discuss how this pretrained model could be finetuned for perturbation response prediction.%
%
\blfootnote{Code available at: \clickableurl{https://github.com/haw-ai-i/SATScG}}
\end{abstract}

\section{Introduction}
\label{sec:intro}

Generating realistic single-cell gene expression vectors for a cell type is broadly useful: such synthetic gene expression vectors can augment scarce or costly measurements, enable in-silico experimentation, and provide a learned representation for downstream analysis. Generating gene expression vectors for a cell line first requires representing that cell line. Existing methods represent a cell line in one of three ways: first, using the categorical name of the cell line, as in CPA~\cite{lotfollahi2023predicting}, GEARS~\cite{roohani2023predicting}, and Lingshu-Cell~\cite{zhang2026lingshucell}; second, using an example gene expression vector from the cell line, as in scGen~\cite{Lotfollahi2019scGen}, scGPT~\cite{cui2024}, Geneformer~\cite{theodoris2023}, PRiMeFlow~\cite{yan2026primeflow}, and scREPA~\cite{wang2026screpa}; or third, using a sequence of gene expression vectors from the cell line, as in State~\cite{adduri2025state}.

In this paper, we propose an autoregressive transformer architecture, paired with a quantized VAE for tokenization, that represents a cell line from a sequence of single-cell gene expression vectors of an arbitrary length, and we study both the fidelity of the generated gene expression vectors and how its pretraining loss scales. The model maps each single-cell gene expression vector to a short sequence of discrete codes, treating discrete codes similarly to how a language model treats tokens. Specifically, our contribution is fivefold: first, we formulate a self-supervised learning task of generating synthetic gene expression vectors given a cell line; second, we define downstream metrics to evaluate the biological fidelity of the generated gene expression vectors; third, we formulate a pretraining loss based on autoregressive cross-entropy prediction over discrete codes; fourth, we show that this pretraining loss correlates with downstream evaluation metrics; and fifth, we study the joint scaling laws of this loss with respect to model size $N$ and pretraining data size $D$. Finally, we discuss how this pretrained cell-line embedding model could be finetuned into a system for perturbation response prediction, and report an observation from pretraining bearing on this idea: a held-out loss on the target perturbation benchmark, tracked throughout pretraining, traces a checkmark-shaped trajectory rather than falling together with the scaling-law loss (Section~\ref{sec:discussion}).

In language and vision, loss as a function of model size $N$ and training data size $D$ follows a predictable power-law form that can be used to choose a compute-optimal allocation between $N$ and $D$~\cite{kaplan2020,hoffmann2022}. Whether an analogous scaling law holds for single-cell foundation models is the subject of an active debate, as discussed in Section~\ref{sec:related}. Several recent papers \cite{kendiukhov2026,wang2026xcell} report power-law scaling along a single axis, such as model size at a fixed data volume or vice versa; at least three independent groups \cite{denadel2026,wang2026xcell,dibaeinia2026} report evidence that single-cell foundation models plateau well short of their largest scales and attribute this to a lack of \emph{context diversity} rather than insufficient capacity; and to our knowledge no prior work has fit a joint, two-exponent $L(N,D)$ law to a single-cell transformer on an independently-varied $N \times D$ grid.

We define the task, metrics, loss, and model in Section~\ref{sec:overview}; show that generation quality tracks the pretraining loss in Section~\ref{sec:gen-result} and present the fitted scaling law with its diagnostics and compute-optimal frontier in Section~\ref{sec:fit-result}; and discuss the implications for the single-cell scaling debate and the Stage 2 perturbation-response direction in Section~\ref{sec:discussion}.

\section{Background and Related Work}
\label{sec:related}

\subsection{Single-cell foundation models}
Single-cell foundation models pretrain on large, heterogeneous atlases and transfer to downstream tasks such as cell-type annotation, perturbation response, or batch correction. Architecturally, they fall into two broad families. \emph{Continuous-embedding} models, such as scVI~\cite{lopez2018}, scGPT~\cite{cui2024}, and the five architectures surveyed in~\cite{denadel2026}, encode expression directly as continuous vectors. In contrast, discretizing models convert expression profiles into sequences of discrete symbols before applying standard sequence models. For example, Geneformer~\cite{theodoris2023} ranks genes by expression and feeds the resulting rank order to a BERT-style encoder. Similarly, Cell2Sentence~\cite{levine2023} and its scaled successor C2S-Scale~\cite{rizvi2026} represent each cell as a sentence of gene names ordered by expression, which is then modeled by a standard language model. Finally, Tahoe-x1~\cite{gandhi2025} bins continuous expression values using a fixed, non-learned quantile codebook. Our model also belongs to this discretizing family; it uses a variational autoencoder with a residual codebook as a tokenizer to learn a discrete representation of individual cell profiles.

\subsection{Neural scaling laws}
\citet{kaplan2020} showed that pretraining loss for transformers trained on natural language follows power laws in model size and data size individually. \citet{hoffmann2022} (Chinchilla) extended this to a joint form and described three ways to estimate it: \emph{Approach 1} fits the lower envelope of training curves; \emph{Approach 2} fits parabolas to loss at fixed compute, representing isoFLOP profiles; \emph{Approach 3} directly fits a parametric two-term loss
\begin{equation}
L(N,D) = E + \frac{A}{N^{\alpha}} + \frac{B}{D^{\beta}}
\label{eq:chinchilla}
\end{equation}
to every $(N,D,L)$ triple by least squares, and inverts the fitted form analytically to obtain a compute-optimal frontier $N^\ast(C)$. We use Approach 3, detailed in Appendix~\ref{S-sec:fit}, and reuse this fitting procedure as-is; our novelty is the application to a discrete-tokenization single-cell transformer, not the fitting procedure itself.

\section{Task, Metrics, and Model}
\label{sec:overview}

This section defines the self-supervised learning task, the metrics that evaluate the generated gene expression vectors, the pretraining loss, and the model architecture; the full specification is deferred to Appendices~\ref{S-sec:method}--\ref{S-sec:exp-design}\ofsupp.

\subsection{The task}
\label{sec:overview-task}

We study a self-supervised learning task: given a set of single-cell gene expression vectors from a cell type, generate additional gene expression vectors of that cell type. Here a cell type corresponds to a group of single-cell gene expression vectors sharing a cell-type label; the label is used only to group vectors into a sentence and is never explicitly included in the model's input. We propose to extract cell type embedding vectors from the pre-trained model for downstream use, as discussed in Section~\ref{sec:discussion}.

\subsection{Metrics}
\label{sec:overview-metrics}

We evaluate a generated population of gene expression vectors against empirical held-out vectors of the same cell type at the level of population averages: each population is summarized by its per-gene mean expression vector, and every metric we report is a comparison between two such average vectors. Metrics of this form are the standard measure of biological fidelity in recent single-cell expression prediction work, including the Virtual Cell Challenge~\cite{roohani2025vcc}. We report three quantities, averaged over cell types: the Pearson correlation between the average vector of the generated population and that of the empirical one, the mean absolute error between the same two average vectors, and a cell-type discriminability score that is high when a generated population's average vector is nearest the empirical average vector of its own cell type among all types in the hold-out set. These metrics are defined in Appendix~\ref{S-sec:gen-eval} (Eqs.~\ref{S-eq:gen-pearson}--\ref{S-eq:gen-disc}); the baseline we score them against is introduced in Section~\ref{sec:gen-result}.

\subsection{The pretraining loss}
\label{sec:overview-loss}

Each single-cell gene expression vector is mapped to a sequence of eight discrete tokens, one per residual quantizer codebook (Section~\ref{sec:overview-model}). The token sub-sequences corresponding to cells from a single cell line are arranged in an arbitrary order to form a sequence, also called a sentence.  The model is trained to autoregressively predict the next token in the sentence. We use the cross-entropy loss, as detailed in Appendix~\ref{S-sec:eval}.

\subsection{Model architecture}
\label{sec:overview-model}

Inspired by approaches in large vision models~\cite{bai2023sequential}, the model has two components, trained in sequence and then composed: a residual quantized variational autoencoder (RQ-VAE)~\cite{lee2022} that learns a discrete tokenizer for single-cell gene expression vectors, and a causal transformer that models sequences of these tokens, as described in Appendix~\ref{S-sec:method}.

\paragraph{Tokenizer.} The tokenizer takes a single-cell log-normalized expression vector over a fixed 18{,}080-gene panel and compresses it to a single 256-dimensional latent vector, which a residual vector quantizer~\cite{lee2022} then encodes coarse-to-fine over eight passes: the first codebook contributes the codeword nearest the latent vector, and each subsequent codebook provides the codeword closest to the residual of the preceding steps. A single-cell expression profile entering as a vector of 18{,}080 expression values is thus mapped to an ordered tuple of eight integers; see Appendix~\ref{S-sec:vqvae} for a concrete example. Because this compression is lossy, it establishes an upper bound on the fidelity with which any cell state can be reconstructed, regardless of the downstream predictor's capacity. The RQ-VAE is trained once and then \emph{frozen}, serving purely as a tokenizer for all subsequent transformer experiments.

\paragraph{Transformer and training.} The tokenizer maps each single-cell gene expression vector to a block of eight discrete tokens, and a set of vectors sharing a cell type is concatenated into a sentence of discrete tokens, one block per vector, separated by a learned separator token. Every sentence holds 32 vectors, or 256 tokens plus the 31 separators, and this length is fixed throughout training and evaluation. Details are provided in Appendix~\ref{S-sec:transformer}. The model builds upon the LLaMA architecture~\cite{touvron2023}, adding a learned positional embedding over the eight tokens encoding a single vector, shared across all vectors in the sequence, together with a learned bias distinguishing tokens that encode a vector from the separators.

\paragraph{Context embedding.} Beyond next-token prediction, the same transformer yields a fixed-size embedding for a cell-line sentence: we mean-pool its last hidden state over the tokens of the cell-line sequence, producing one vector for the context, as detailed in Appendix~\ref{S-sec:transformer}. \anonalt{Averaging token-level hidden states over the sequence dimension is the standard way to read a sequence-level embedding out of a decoder-only model; here that operation is applied to a cell-line context rather than to text.}{This mirrors the mechanism REAL~\cite{zhang2024real} uses to extract response embeddings from decoder-only language models, averaging token-level hidden states over the sequence dimension; here the same operation is applied to a cell-line context rather than a dialogue response.} For downstream perturbation response prediction (Section~\ref{sec:discussion}), this pooled vector serves as the cell-line context embedding, giving predictors an alternative to the categorical, single-profile, or full-sentence cell-line representations surveyed in Section~\ref{sec:intro}.

\section{Results}
\label{sec:results}

\subsection{Gene expression vector generation quality tracks the pretraining loss}
\label{sec:gen-result}
\begin{figure*}[t]
\centering
\includegraphics[width=0.95\textwidth]{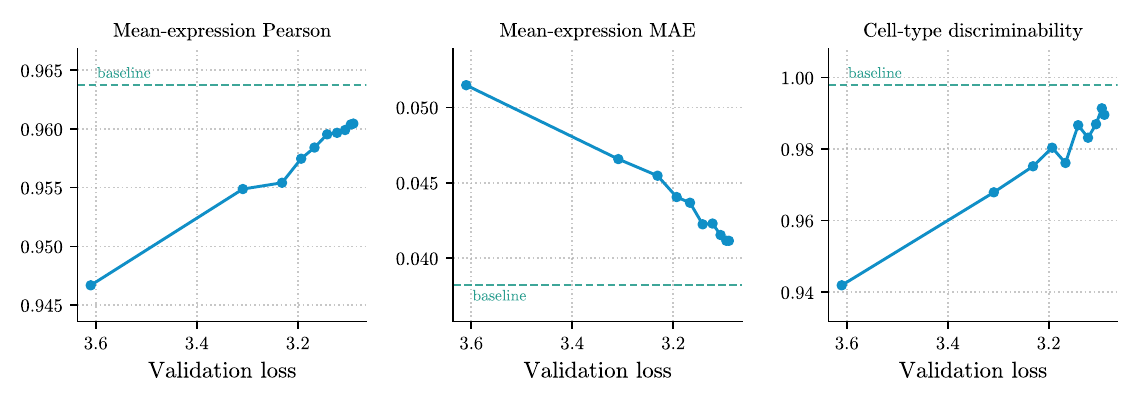}
\caption{Evaluation of single-cell gene expression vector generation quality metrics. Each
panel plots a generation-quality metric for the $4.43$M model against its validation loss across ten training checkpoints (loss falls left to right, so training progresses rightward). The dashed horizontal line marks the baseline (Table~\ref{tab:generation}). The model tracks close to it, improving as the loss decreases.}
\label{fig:gen-trends}
\end{figure*}

We first ask whether the cross-entropy loss is predictive of the target evaluation metrics of the task, verifying that lower pretraining loss yields higher gene expression vector generation quality. For this evaluation, we measure generation quality across ten training checkpoints of a $4.43$M-parameter model, during which the validation loss decreases from $3.61$ to $3.09$. We score the model against a \emph{baseline}: the value each metric takes when a second, disjoint set of empirical held-out vectors of the same cell type, passed through the same tokenizer as the generated ones, is used in place of the generated vectors. It is the score a near-ideal generator would obtain. As detailed in Appendix~\ref{S-sec:gen-eval}\ofsupp, the generation quality improves monotonically as the loss falls and tracks it closely. The mean-expression Pearson correlation coefficient rises from $0.947$ to $0.960$, exhibiting a Pearson correlation $r=-0.99$ against validation loss across checkpoints. Concurrently, the mean-expression mean absolute error decreases from $0.051$ to $0.041$, with a correlation $r=+0.98$. These trends are illustrated in Figure~\ref{fig:gen-trends}, with the per-checkpoint values listed in Table~\ref{S-tab:gen-trend}. Table~\ref{tab:generation} reports the lowest-validation-loss checkpoint against the baseline.


\begin{table}[t]
\centering
\begin{tabular}{@{}lcc@{}}
\toprule
\toprule
Metric & Model & Baseline \\
\midrule
Mean-expression Pearson ($\uparrow$)   & 0.960 & 0.964 \\
Mean-expression MAE ($\downarrow$)     & 0.041 & 0.038 \\
Cell-type discriminability ($\uparrow$) & 0.990 & 0.998 \\
\bottomrule
\end{tabular}
\caption{Same-cell-type generation quality for the $4.43$M model at the checkpoint with the lowest pretraining validation loss (per-checkpoint values in Table~\ref{S-tab:gen-trend}), scored against held-out empirical vectors and averaged over $131$ cell types. Metrics are defined in Eqs.~\ref{S-eq:gen-pearson}--\ref{S-eq:gen-disc}.}
\label{tab:generation}
\end{table}

First, generation is strongly cell-type-specific. Discriminability is $0.990$, essentially at the baseline of $0.998$ and far above the $\approx 0.5$ a random ordering would give: a generated population's mean vector is almost always closest to the empirical mean vector of its own cell type among all $131$ types. The generated gene expression vectors are therefore about as cell-type-discriminable as empirical biological replicates.

Second, the generated gene expression vectors match empirical replicates on this measure. Their mean-expression Pearson ($0.960$) and MAE ($0.041$) sit just under the baseline ($0.964$, $0.038$), meaning the generated population matches empirical vectors almost as closely as two empirical subsets of the same cell type match each other. This baseline falls short of a perfect score of $1.0$ for correlation or $0.0$ for error because of the lossy tokenization noted in Section~\ref{sec:overview-model} and because the metric compares two finite, distinct biological samples. This is a within-run, single-model-size probe rather than a downstream scaling result, as discussed in Appendix~\ref{S-sec:limitations}, but it shows that reductions in the pretraining loss translate into measurable, cell-type-specific generation improvements. Generation this faithful also points to a practical use: the model can serve as a generator of synthetic single-cell gene expression vectors, for example to augment scarce measurements or to produce in-silico populations of a given cell type.

\subsection{The single-cell scaling debate}
\label{sec:related-table}
Having seen that generation quality tracks the pretraining loss, we now turn to how that loss scales with {model size $N$ and data size $D$}. Whether single-cell foundation models scale predictably is actively contested. At least six single-cell papers posted in the H2 of 2025 and H1 of 2026 directly address model- or data-scaling for single-cell foundation models; Table~\ref{tab:related} summarizes how each relates to our design. None reports an independently-varied $N \times D$ grid fit with a joint two-exponent law of the form introduced by \citet{hoffmann2022}, and a compute-optimal frontier: prior work either sweeps a single axis, co-varies $N$ and $D$, or reports only a one-exponent $N$-only power law.

\begin{table*}[t]
\centering
\small
\setlength{\tabcolsep}{1mm}
\begin{tabular}{@{}p{3.6cm}p{4.2cm}p{1.4cm}p{3.0cm}p{3.6cm}@{}}
\toprule
Work & $N$ range (\# sizes) & Joint $N{\times}D$? & Parametric fit? & Perturbation-task scaling? \\
\midrule
Geneformer scaling~\cite{geneformer-scaling} & 38M--316M (5), fixed $D$ & No & No & Not studied \\
C2S-Scale~\cite{rizvi2026} & 410M--27B (4), fixed $D$ & No & No & 2 of 4 sizes only \\
Tahoe-x1~\cite{gandhi2025} & 70M--3B (3), $D$ confounded with $N$ & No & No & Frozen embeddings + separate model \\
Masked-recon.\ transformers~\cite{kendiukhov2026} & 533--100.5M (6), fixed $D{=}200$k & No (single-axis) & Power law ($N$-only) & Not studied \\
X-Cell~\cite{wang2026xcell} & 83M--3.1B (5), fixed $D$ & No & Power law ($N$-only) & Yes, training loss only \\
\citet{denadel2026} & 5 architectures, $N$ minimally varied & No & No & Not studied \\
\textbf{Ours} & 1.3M--83.9M (5, fixed ratio) & \textbf{Yes} & \textbf{Chinchilla Approach 3} & Pretraining loss \\
\bottomrule
\end{tabular}
\caption{Scaling-oriented single-cell papers most relevant to this work. \emph{Confounded} means model size co-varies with another factor, such as data, architecture family, or an extra input token, rather than being swept on an otherwise-fixed setup.}
\label{tab:related}
\end{table*}

This design difference changes the conclusion. Three studies in this literature, which we refer to throughout as the \emph{diversity-over-scale} cluster~\cite{denadel2026,wang2026xcell,dibaeinia2026}, report no clear data-scaling law, or saturation well below the scales they test, and read this as evidence that scale is not the primary lever. Under the two-term parametric loss of \citet{hoffmann2022} in Eq.~\ref{eq:chinchilla}, however, increasing $D$ at \emph{fixed or uncontrolled} $N$ is \emph{expected} to plateau once the data term $B/D^{\beta}$ falls below the capacity term $A/N^{\alpha}$, since the model then lacks the capacity to exploit more data. That is a capacity bottleneck, rather than proof that a data-scaling law does not exist, and distinguishing the two requires sweeping $N$ and $D$ independently, which none of these studies do. Our grid does, and on it the loss does scale: both exponents come out clearly positive, the fitted law predicts runs held out of it, and inverting it gives a compute-optimal frontier (Section~\ref{sec:fit-result}). Appendix~\ref{S-sec:scaling-debate} treats the three studies and this argument in detail, and Section~\ref{sec:discussion} returns to it in light of our results.

\subsection{A controlled grid varies model size and data independently}
\label{sec:overview-grid}

We therefore construct an explicit grid that varies $N$ and $D$ independently on an otherwise fixed architecture family and data composition, as detailed in Appendix~\ref{S-sec:exp-design}.

Two controls make the axes clean. On the model axis, a naive sweep over parameter count alone confounds size with shape, since many width/depth pairs give the same parameter count; we instead hold the shape of the network fixed across sizes, scaling width and depth together in fixed proportion, and verify each parameter count by instantiating the model. Five configurations span a factor of $\approx64\times$ in parameter count, from $1.31$M to $83.9$M. On the data axis, $D$ is the number of single-cell gene expression vectors the model is trained on, representing the single-cell analogue of counting training tokens, sampled from a fixed $\approx200$M-vector subset of the Arc Institute scBaseCount atlas while holding its composition, representing the relative mix of tissues, studies, and cell types, fixed. This sampling strategy ensures that $D$ isolates the effect of data quantity rather than diversity, as described in Appendix~\ref{S-sec:data}. This gives us 22 distinct $(N,D)$ points, with $D$ running from $38.1$M to $362$M gene expression vectors, and up to $1.01$B on the held-out runs of Section~\ref{sec:heldout}.

Runs do not cycle repeatedly through the corpus, so the held-out loss we fit coincides with the training loss. Within each cell type, $20\%$ of cells are reserved as a validation split that no run ever trains on, and $L(N,D)$ is the pretraining cross-entropy measured on that split. Because $D$ counts cells processed against a training pool of $\approx160$M cells, a run makes at most a single pass over that pool at most grid points, with repetition only at the largest budgets, leaving little opportunity to memorize. On a representative run the validation and training losses end at $3.09$ and $3.10$, a negligible generalization gap. Appendix~\ref{S-sec:data} gives the split and the sampling scheme, and Appendix~\ref{S-app:sampling} shows the full training and validation curves.

This is deliberately a small-scale setup: the largest model in the grid has at most $83.9$M parameters, two orders of magnitude below the billion-parameter models in Table~\ref{tab:related}, and all experiments ran on a single NVIDIA H100 GPU. Reproducing the scaling law therefore does not require large-scale infrastructure, and should be within reach of teams with modest compute budgets.

\subsection{A two-exponent scaling law for the pretraining loss}
\label{sec:fit-result}
We fit a controlled scaling law over the 22 $(N,D)$ points of our grid, fitting Eq.~\ref{eq:chinchilla} as described in Appendix~\ref{S-sec:exp-design}. On the self-supervised pretraining loss, this yields a two-term power law of the standard Chinchilla form, with both exponents clearly positive:
\begin{equation}
L(N,D) \;=\; \underbrace{2.90}_{E} \;+\; \frac{2.05\times 10^{4}}{N^{0.81}} \;+\; \frac{4.28\times 10^{3}}{D^{0.57}},
\label{eq:fitted}
\end{equation}
i.e.\ $\alpha \approx 0.81$ (returns to model size) and $\beta \approx 0.57$ (returns to data), with an irreducible-loss floor of $E \approx 2.90$ nats. The fit explains the grid well: $R^2 = 0.977$ on the loss and $0.976$ on the log-loss, with a log-loss RMSE of $0.0046$ and a mean absolute error of $0.008$ nats. The residuals, shown in Figure~\ref{S-fig:residuals} and detailed in Appendix~\ref{S-app:residuals}, are small, roughly symmetric about zero, and show no systematic trend with $N$ or $D$; the largest single residual is under $0.02$ in log space, representing an error of less than 2\%.

Because $\alpha > \beta$, returns to model capacity fall off faster than returns to data over this regime, and inverting the law gives a compute-optimal allocation, Eq.~\ref{S-eq:frontier}, that tilts toward data: $D^\ast \propto C^{\alpha/(\alpha+\beta)} \approx C^{0.59}$ against $N^\ast \propto C^{0.41}$, so at these scales each additional unit of compute is better spent on more cells than on more parameters. Recovering a data exponent this clearly positive is what the $\approx200$M-cell corpus of Appendix~\ref{S-sec:data} buys, since a grid that varies $D$ over a narrow range, or only at fixed $N$, leaves the data term of Eq.~\ref{eq:chinchilla} too flat to constrain $\beta$ (Section~\ref{sec:related-table}).

\begin{figure*}[t]
\centering
\includegraphics[width=0.75\textwidth]{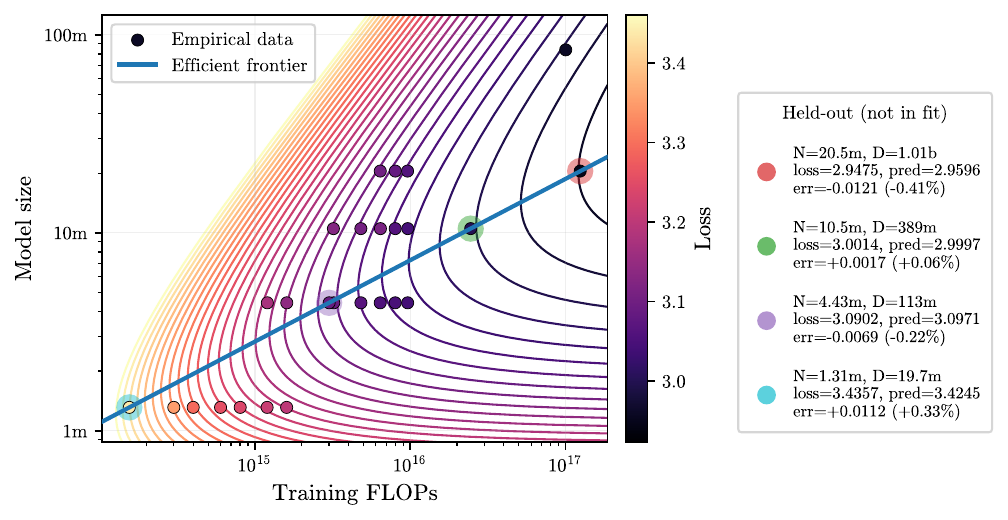}
\caption{Fitted scaling surface for the self-supervised pretraining loss. Iso-loss contours of Eq.~\ref{eq:fitted} are shown in the $(\text{model size},\ \text{training FLOPs})$ plane with $C\approx 6ND$; filled circles are the 22 empirical grid points, colored by measured loss; the solid ascending line is the compute-optimal frontier $N^\ast(C)$ of Eq.~\ref{S-eq:frontier}. The four labeled points, with measured and predicted losses in the legend, are runs held \emph{out} of the fit, one at each model size below $83.9$M; we discuss them in Section~\ref{sec:heldout}.}
\label{fig:isoloss}
\end{figure*}

\subsection{The fitted law predicts held-out runs within 0.5\%}
\label{sec:heldout}
A parametric fit with five free parameters on 22 points can, in principle, absorb structure that would not generalize. To guard against this, we held four runs, one at each of the four smaller model sizes, completely out of the fit and evaluated the ability of the fitted law to predict their losses, representing the labeled points in Figure~\ref{fig:isoloss}. Three of the four fall outside the data budgets the fit sees at their model size, so predicting them tests extrapolation along $D$; the fourth tests interpolation. The fitted law predicts all four held-out losses within $0.5\%$ of the measured values. For $N=1.31$M at $D=19.7$M, the predicted loss of $3.4245$ is within $+0.33\%$ of the $3.4357$ measured. For $N=4.43$M at $D=113$M, the predicted loss of $3.0971$ is within $-0.22\%$ of the $3.0902$ measured. For $N=10.5$M at $D=389$M, the predicted loss of $2.9997$ is within $+0.06\%$ of the $3.0014$ measured. Finally, for $N=20.5$M at $D=1.01$B, which is the largest data budget in the study sitting on the compute-optimal frontier, the predicted loss of $2.9596$ is within $-0.41\%$ of the $2.9475$ measured. That the law predicts held-out $(N,D)$ points this closely, across every model size, gives us more confidence in reading the compute-optimal frontier of Eq.~\ref{eq:fitted} than the in-sample $R^2$ alone would warrant. Every run in the grid ends where its data budget runs out, not where the model starts to memorize: the validation loss tracks the training loss throughout and is still at its minimum at the last measured step. We can therefore read the fitted $\beta$ as a return to data. This comes from the proportional pool sampling of Appendix~\ref{S-sec:data} rather than from the size of the corpus, since uniform pool weighting on the same data oversamples the rare cell types and turns the validation loss upward part-way through a run (Appendix~\ref{S-app:sampling}).
\section{Discussion}
\label{sec:discussion}

\paragraph{A scaling law exists for this architecture: on the pretraining
objective.} Our fit shows that the self-supervised pretraining loss of an RQ-VAE-tokenized single-cell transformer follows a joint two-exponent law over the range we test, and predicts held-out points well enough, as shown in Section~\ref{sec:heldout}, to determine a compute-optimal allocation. This yields a positive scaling result for a discrete, learned-tokenization architecture, a family not previously fit with a joint $N\times D$ law, as summarized in Table~\ref{tab:related}. This also helps explain the no-data-scaling plateaus reported elsewhere. Those studies mostly add data at a fixed or uncontrolled model size, and under the fitted law, this is precisely when additional data ceases to reduce the loss. Once the data term $B/D^{\beta}$ falls below the capacity term $A/N^{\alpha}$, the model lacks the capacity to leverage more data. The loss consequently flattens even though more data is available. This behavior can be misconstrued as a limit to data scaling, whereas it represents a model-size bottleneck, supporting the argument in Section~\ref{sec:related-table}. A fixed-$N$ slice through our fitted surface exhibits the same plateau, even though scaling $N$ and $D$ together consistently reduces the loss.

\paragraph{Scope: this is a statement about the pretraining objective.} The
diversity-over-scale papers~\cite{denadel2026,wang2026xcell,dibaeinia2026} make their strongest claims about downstream biological signal, such as differentially expressed gene recovery or perturbation discrimination, rather than the pretraining loss. Our result does not contradict theirs: a well-determined scaling law for the pretraining loss can hold at the same time as a weak downstream trend, if what limits the downstream tasks is the diversity of the available data rather than model size or the amount of pretraining. What our joint grid adds is a controlled demonstration that, for the self-supervised objective, the apparent lack of data scaling in single-axis studies arises from a model-size bottleneck rather than being an inherent property of single-cell transformers. The two-term law appears clearly once $N$ and $D$ are varied together over a sufficiently wide range. As a first step toward addressing downstream tasks, Section~\ref{sec:gen-result} demonstrates that same-cell-type generation quality tracks the pretraining loss, producing synthetic gene expression vectors that are comparable to empirical replicates on a gene-expression-profile measure.

\paragraph{Stage 2: an idea for perturbation-conditioned generation, and an observation from pretraining.} The pretraining objective studied above (Stage 1) builds a general-purpose cell-line representation. The way we envision turning it into a perturbation-response predictor is a second stage that finetunes this pretrained model on a perturbation-conditioned objective: a block of control-vector tokens, a token identifying the perturbation, and a block of perturbed-vector tokens, with the model scored on predicting the perturbed vector's tokens from the control cell-line context and the perturbation identity. This pretrain-then-finetune (Stage 1 $\to$ Stage 2) framework is the direction we are pursuing to apply the architecture to perturbation response; we do not report Stage 2 results here. What we do report is an observation from Stage 1 pretraining bearing on this idea. While monitoring Stage 1 pretraining, we track a held-out loss on control single-cell gene expression vectors drawn from the same perturbation benchmark a Stage 2 finetune would target, scored in the ordinary (unconditioned) pretraining format. This loss does not fall together with the scBaseCount loss the scaling law is fit to: it drops sharply over the first few hundred steps and then rises for the remainder of the run, tracing a checkmark-shaped trajectory (Appendix~\ref{S-app:checkmark}), even as the scBaseCount validation loss keeps falling over the same steps. A natural reading is that the two losses diverge due to a composition mismatch: scBaseCount's cross-tissue, cross-study composition increasingly dominates the representation as pretraining proceeds, at the expense of fit to the narrower, cell-line-specific slice the perturbation benchmark occupies. This raises an open question for the Stage 1 $\to$ Stage 2 framework, namely whether and how the point in Stage 1 pretraining at which a representation is handed off matters for how well it finetunes for perturbation response, which we leave to future work.
\section{Conclusion}
\label{sec:conclusion}

We presented an RQ-VAE-tokenized, LLaMA-architecture transformer for single-cell gene expression modeling, pretrained self-supervised on the scBaseCount atlas to synthesize expression vectors given a cell-line context, and evaluated two primary properties. First, its generation quality: conditioned on held-out vectors of a cell type, it generates cell-type-specific gene expression vectors that match the biological fidelity of empirical replicates on a gene-expression-profile measure, and this quality tracks the pretraining loss. Second, how that loss scales: on a controlled grid that varies model size $N$ and data budget $D$ independently, spanning $1.3$--$83.9$M parameters and data budgets of $38.1$M--$362$M vectors on a $\approx200$M-vector corpus, we fit the Chinchilla Approach-3 parametric loss and obtain, to our knowledge, the first jointly-fit two-exponent scaling law and compute-optimal frontier for a single-cell transformer, with $\alpha\approx0.81$ and $\beta\approx0.57$, $R^2\approx0.98$, and out-of-sample predictions that track measured losses closely. This locates the no-data-scaling plateaus reported elsewhere as a downstream, not an upstream, phenomenon. Applying the pretrained model to perturbation response, through the Stage 2 perturbation-conditioned generation the architecture is designed for, is the natural direction this result opens up.

\section*{Acknowledgments}

We thank Rong Wei, Alexander Avdoshkin, Ziyi Zhuang, and Ziyu Chen for their contributions to this work, and Leonid Klarov for providing the compute infrastructure that supported the GPU experiments reported here.

This work was supported by the National Science Foundation NRT-AI 2244574 and through allocation number CIS240027 from the Advanced Cyberinfrastructure Coordination Ecosystem: Services \& Support (ACCESS) program, which is supported by National Science Foundation grants \# 2138259, \# 2138286, \# 2138307, \# 2137603, and \# 2138296.

The technical support and advanced computing resources from University of Hawaii Information Technology Services - Research Cyberinfrastructure, funded in part by the National Science Foundation CC* awards \# 2201428 and \# 2232862 are gratefully acknowledged.

This work was supported by computational resources provided by NPC Labs through B3-IQ infrastructure platform.


\bibliography{references}

\appendix

\section{Method}
\label{sec:method}

Our model has two components, trained in sequence and then composed: (i) a RQ-VAE that learns a discrete tokenizer for single-cell expression profiles, and (ii) a causal transformer that models sequences of these tokens.

\subsection{RQ-VAE tokenizer}
\label{sec:vqvae}

Each cell is represented as a vector $x \in \mathbb{R}^{18{,}080}$ of log-normalized expression over a fixed gene panel (Section~\ref{sec:data}). An MLP encoder $f_\theta$ maps $x$ to a latent $z_e = f_\theta(x) \in \mathbb{R}^{256}$ (one hidden layer of width 1{,}024 with a ReLU nonlinearity in our default configuration); $z_e$ is quantized by a residual vector-quantizer~\cite{lee2022} with 8 residual codebooks of 256 entries each (initialized via k-means on the first batch), giving each cell a sequence of 8 discrete tokens $t_1,\dots,t_8 \in \{0,\dots,255\}$; an MLP decoder $g_\theta$ reconstructs $\hat{x} = g_\theta(z_q)$ from the quantized latent $z_q$. At step $k$, codebook $C^{(k)}\in\mathbb{R}^{256\times256}$ selects $C^{(k)}_{t_k}$; the residual sequence is defined by $r^{(1)}=z_e$ and $r^{(k+1)}=r^{(k)}-C^{(k)}_{t_k}$. The training objective is
\begin{equation}
\mathcal{L}_{\text{RQ-VAE}} = \underbrace{\lVert \hat{x} - x \rVert_2^2}_{\text{reconstruction}} \;+\; \lambda_c \sum_{k=1}^{8} \lVert r^{(k)} - \mathrm{sg}[C^{(k)}_{t_k}] \rVert_2^2,
\end{equation}
the sum of an MSE reconstruction loss and a commitment loss at each residual codebook $k$ (stop-gradient $\mathrm{sg}[\cdot]$, commitment weight $\lambda_c = 0.25$), summed (not averaged) across codebooks. We train with AdamW, a linear warmup (1{,}000 steps, or 10\% of total steps for short runs) followed by cosine decay to a minimum learning rate of $10^{-6}$, and batch size 1{,}024. The RQ-VAE is trained on the scBaseCount atlas; the resulting checkpoint is then \emph{frozen} and used purely as a tokenizer for all transformer experiments in Section~\ref{sec:exp-design}.

\paragraph{Worked example: from expression to eight codes.} It is worth
making the expression-to-token map concrete, since it is the step that turns a continuous cell into a short string of symbols. A single cell enters as its log-normalized expression vector and leaves as an ordered tuple of eight integers,
\begin{equation*}
\begin{aligned}
x \in \mathbb{R}^{18{,}080}
  &\;\xrightarrow{\;f_\theta\;}\; z_e \in \mathbb{R}^{256}\\
  &\;\xrightarrow{\;\text{ResidualVQ}\;}\; (t_1,\dots,t_8) \in \{0,\dots,255\}^8\\
  &\;\xrightarrow{\;g_\theta\;}\; \hat{x} \in \mathbb{R}^{18{,}080}.
\end{aligned}
\end{equation*}

The encoder $f_\theta$ compresses the $18{,}080$-dimensional profile to a single $256$-dimensional vector $z_e$, and the residual quantizer then encodes $z_e$ \emph{coarse-to-fine} over eight passes. Using the residual sequence defined above, the token selected from each codebook is
\begin{equation}
t_k = \operatorname*{arg\,min}_{j\in\{0,\dots,255\}}
       \bigl\lVert r^{(k)} - C^{(k)}_j \bigr\rVert_2,
\qquad k = 1,\dots,8,
\end{equation}
so $t_1$ names the codeword closest to $z_e$ itself, $t_2$ names the codeword that best explains what codebook~1 left over, and so on down to $t_8$; the quantized latent handed to the decoder is the running sum $z_q = \sum_{k=1}^{8} C^{(k)}_{t_k}$. Concretely, a T~cell whose profile is dominated by markers such as \emph{CD3D} and \emph{IL7R} might encode as
\[
\begin{gathered}
z_e \;\longmapsto\; (t_1,\dots,t_8)\\
 = (46,\,201,\,46,\,133,\,8,\,224,\,79,\,150),
\end{gathered}
\]
with illustrative values, so the entire cell is now represented by the eight-symbol string $46\;\,201\;\,46\;\,133\;\,8\;\,224\;\,79\;\,150$, which is the block of tokens that this cell contributes to a cell sentence (Section~\ref{sec:transformer}). Although the eight codebooks are distinct, their indices share the single range $\{0,\dots,255\}$, so the transformer reuses one $256$-row embedding table across all codebooks: the codes drawn from codebooks~1 and~3 above are both written $46$ and map to the same embedding row, and only the within-cell position embedding (Section~\ref{sec:transformer}) distinguishes them.

\subsection{Cell-sentence transformer}
\label{sec:transformer}

We reuse the LLaMA decoder-only architecture~\cite{touvron2023} (via HuggingFace's \texttt{LlamaConfig}/\texttt{LlamaForCausalLM}), trained from scratch rather than from pretrained language-model weights, as the backbone for modeling sequences of RQ-VAE tokens. The vocabulary is the union of the 256 RQ-VAE codes (shared across all 8 residual codebooks) and a single separator token \texttt{[SEP]}.

\paragraph{Sequence construction.} A pretraining example is a sequence of 32
cells of the same cell type, sampled from the same cell-type pool (cells are grouped by their \texttt{cell\_type} annotation when the pretraining tokens are prepared, and only cell types with at least 1{,}000 cells are kept), each contributing its 8 RQ-VAE tokens, with \texttt{[SEP]} inserted between (but not after) consecutive cells, giving $32\times 8 + 31 = 287$ tokens. Note the cell-type label is used only to decide which cells may share a sequence; it is not fed to the model as a token. The model is trained with standard next-token cross-entropy; loss on \texttt{[SEP]} positions is masked out of the target so the model is only ever scored on predicting RQ-VAE code positions.

\paragraph{Cell-aware position encoding.} On top of the backbone's own position
encoding, we add two signals that mark the block structure of a cell sentence: (a) a learned positional embedding over the 8 within-cell token positions only, shared across all cells in the sequence, so that a token's role within its cell is encoded independently of where that cell falls in the sentence, and (b) one of two learned bias vectors added to every token's embedding depending on whether it belongs to a cell (\emph{intra}) or is a separator or other special token (\emph{inter}).

\paragraph{Cell-line context embedding.} Beyond the per-cell pooling used above, the same last hidden states support a coarser, sequence-level embedding representing an entire cell-line context. Let $h_1,\dots,h_T \in \mathbb{R}^d$ denote the transformer's last hidden states at the $T$ non-separator token positions of a context-cell sequence, excluding \texttt{[SEP]} positions, consistent with how separator positions are already excluded from the pretraining loss target (Section~\ref{sec:pretraining}). The cell-line embedding is the mean of these hidden states,
\begin{equation}
e = \frac{1}{T}\sum_{t=1}^{T} h_t,
\end{equation}
\anonalt{which is the standard average-token-pooling used to obtain sequence-level embeddings from decoder-only models, applied here to a cell-line context}{the same average-token-pooling mechanism REAL~\cite{zhang2024real} uses to obtain sequence-level response embeddings from decoder-only language models, applied here to a cell-line context rather than a dialogue response}. Because every context cell contributes a fixed 8 tokens, $e$ is equivalently the average of the per-cell pooled vectors defined above, taken over every cell in the sequence, so the two poolings agree wherever both are computed on the same sequence. This pooling rule is defined identically regardless of how many cells the context holds, giving a fixed-$d$-dimensional cell-line embedding at any context length (Section~\ref{M-sec:overview-model}).

\subsection{Pretraining}
\label{sec:pretraining}
All experiments in this paper train the model self-supervised by next-token prediction on control-only cell sentences drawn from scBaseCount (Section~\ref{sec:data}). This pretraining objective --- and its scaling with model and data size (Section~\ref{M-sec:results}) --- is the sole training procedure we study.

\subsection{Evaluation metric}
\label{sec:eval}
The quantity we fit is the self-supervised pretraining loss: next-token cross-entropy over held-out scBaseCount cell sentences, measuring how well the model predicts the RQ-VAE code positions of unseen cells. This is the loss $L(N,D)$ in Eq.~\ref{M-eq:chinchilla} and the target of every fit in Section~\ref{M-sec:results}. Because the sequence includes \texttt{[SEP]} positions whose targets are masked out, the loss is averaged only over predicted code positions, so it is comparable across the different model and data sizes of the grid.

\subsection{Same-cell-type generation and its evaluation}
\label{sec:gen-eval}
Beyond the pretraining loss, we ask whether the pretrained model generates realistic gene expression vectors and whether generation quality improves as that loss falls. Using the pretraining (same-cell-type) format, we prompt the model with a set of held-out vectors of a given cell type and autoregressively sample new vectors of that type --- each of the eight code positions in turn, restricted to the RQ-VAE codes --- then decode the generated tokens to expression with the frozen RQ-VAE. For each of $131$ cell types we compare the generated population against a disjoint set of held-out \emph{empirical} vectors of the same type, using their original (log-normalized) expression as the reference. Because a generated vector can only be produced by decoding its RQ-VAE tokens back to expression, any tokenizer reconstruction error falls entirely on the generated side of the comparison; our baseline (defined below) measures exactly that error, so the gap between the model and the baseline isolates the transformer's own contribution. We summarize each comparison with three quantities, averaged over cell types: (i) the Pearson correlation between the per-gene mean expression of generated and empirical vectors (Eq.~\ref{eq:gen-pearson}; profile agreement --- whether the relative pattern across genes matches); (ii) the mean absolute error of that per-gene mean (Eq.~\ref{eq:gen-mae}; magnitude error --- whether the absolute expression levels match); and (iii) a cell-type \emph{discriminability} score in $[0,1]$ (Eq.~\ref{eq:gen-disc}) --- $1$ when each generated population's mean vector is nearest the empirical mean vector of its own cell type among all types, and $\approx 0.5$ under random guessing --- which measures whether generation is cell-type-specific rather than mode-collapsed across types.

Formally, write $C=131$ for the number of cell types and $G=18{,}080$ for the number of genes. For cell type $c$ let $\bar{\mathbf{x}}_c\in\mathbb{R}^G$ be the per-gene mean expression of the generated vectors of that type and $\bar{\mathbf{y}}_c\in\mathbb{R}^G$ that of the disjoint set of empirical hold-out vectors, with $\bar x_{c,g}$, $\bar y_{c,g}$ their $g$-th entries and $\langle\bar{\mathbf{x}}_c\rangle=\tfrac1G\sum_g \bar x_{c,g}$ the average across genes. Writing $u_{c,g}=\bar x_{c,g}-\langle\bar{\mathbf{x}}_c\rangle$ and $v_{c,g}=\bar y_{c,g}-\langle\bar{\mathbf{y}}_c\rangle$ for the gene-centered deviations, the three metrics are
\begin{align}
\mathrm{Pearson} &= \frac{1}{C}\sum_{c=1}^{C} \frac{\sum_{g} u_{c,g}\, v_{c,g}} {\sqrt{\sum_{g} u_{c,g}^2}\; \sqrt{\sum_{g} v_{c,g}^2}},
     \label{eq:gen-pearson}\\[4pt]
\mathrm{MAE} &= \frac{1}{C}\sum_{c=1}^{C}\frac{1}{G}\sum_{g=1}^{G} \bigl\lvert \bar x_{c,g}-\bar y_{c,g}\bigr\rvert,
     \label{eq:gen-mae}\\[4pt]
\mathrm{Disc} &= \frac{1}{C}\sum_{c=1}^{C}\Bigl(1-\frac{\rho(c)}{C}\Bigr),
     \label{eq:gen-disc}\\[4pt]
\rho(c) &= \Bigl\lvert\bigl\{\,c':\; \lVert\bar{\mathbf{x}}_c-\bar{\mathbf{y}}_{c'}\rVert_1 < \lVert\bar{\mathbf{x}}_c-\bar{\mathbf{y}}_{c}\rVert_1\,\bigr\}\Bigr\rvert,
     \nonumber
\end{align}
where $\rho(c)\in\{0,\dots,C-1\}$ is the rank of the true cell type's empirical mean vector when the $C$ empirical mean vectors are ordered by increasing $\ell_1$ distance to $\bar{\mathbf{x}}_c$ (rank $0$ = nearest). Discriminability is therefore $1$ when every generated mean vector is closest to its own type's empirical mean vector and averages $\approx 0.5$ under random ordering. To show how good a prediction can look on each metric, we compare it against a \emph{baseline}, which replaces the generated vectors with a second, disjoint set of empirical held-out vectors of the same type, passed through the same RQ-VAE the generated vectors use. It is the best score attainable by any generator whose outputs must go through the tokenizer, and it still falls short of a perfect score for two reasons: it compares two \emph{different} finite samples of empirical vectors (sampling and biological variation), and the baseline vectors carry the same RQ-VAE reconstruction error as the generated ones.

\section{Data}
\label{sec:data}

\paragraph{Gene panel.} All data is aligned to the fixed 18{,}080-gene
panel used as the interface for the tokenizer (Section~\ref{M-sec:overview-model}); any external dataset used in this work is subset and reordered to this panel before further processing. All expression matrices are normalized to 20{,}000 total counts per cell and log1p-transformed.

\paragraph{Pretraining corpus (scBaseCount).} Pretraining uses the Arc Institute's scBaseCount atlas, downloaded from the public \texttt{arc-institute-virtual-cell-atlas} bucket (human). The full atlas holds on the order of half a billion cells; we work with a fixed subset of $\approx 200$ million cells (23{,}846 source \texttt{h5ad} files after gene alignment and filtering). Cells are grouped by their cell-type annotation (keeping only cell types with at least 1{,}000 cells), and within each cell type $20\%$ of cells are held out for validation. This single corpus, at fixed composition, is the only data source in the scaling grid. The data axis $D$ is the pretraining \emph{data budget}, which is the number of cells the model is trained on, sampled from the corpus (with repetition at the largest budgets). This is the single-cell analogue of counting training tokens in language-model scaling laws, and the quantity that enters the compute proxy $C\approx 6ND$ (Section~\ref{sec:exp-design}); each optimizer step consumes $256\times 32 \approx 33$k cells, so a run's $D$ determines its step budget. Varying $D$ over a wide range at each model size, on this fixed corpus, is what lets us fit the two-term law of Eq.~\ref{M-eq:chinchilla} cleanly.

\paragraph{Sampling the corpus.} Each pretraining sentence is built by first choosing a cell-type pool and then drawing that sentence's $32$ cells from it (Section~\ref{sec:transformer}). We choose a pool with probability \emph{proportional} to the number of cells it holds, so that every cell in the corpus is equally likely to be trained on regardless of which cell type it belongs to, and the corpus's natural, long-tailed composition over cell types is preserved in expectation, yielding the fixed composition that the scaling grid holds constant while $D$ varies. The obvious alternative, weighting every cell type equally, instead rebalances training toward the rare types. Because a rare type's pool is small (a type is kept only if it holds at least $1{,}000$ cells, while the common types contribute orders of magnitude more), its few cells are revisited many times over a run, and the model memorizes them: the training loss keeps falling while the validation loss stops tracking it and turns upward part-way through training. Appendix~\ref{app:sampling} shows this failure directly. Every run reported in this paper uses proportional sampling.

\section{The Single-Cell Scaling Debate in Detail}
\label{sec:scaling-debate}

This appendix expands on the comparison summarized in Table~\ref{M-tab:related} and Section~\ref{M-sec:related-table}, describing the three studies of the \emph{diversity-over-scale} cluster and the structural argument that separates a capacity bottleneck from a missing data-scaling law.

Three of these results converge on a similar conclusion from different angles. \citet{denadel2026} pretrain 400 models across 6{,}400 experiments on a 22.2M-cell corpus and report that single-cell foundation models plateau in performance with pretraining datasets that are only a fraction of the size of current training corpora, with no clear data-scaling law. \citet{wang2026xcell}'s own appendix reports that biological signal, specifically the differentially expressed gene Pearson correlation, for their largest model family saturates around 1.6B parameters, well below their 3.1B ceiling, and attributes this to dataset diversity, with approximately 10{,}700 unique perturbation-context sets, rather than model capacity. \citet{dibaeinia2026} is explicitly a position paper arguing against scaling as the primary lever, and provides controlled evidence on a fixed architecture without varying model size that at matched cell counts, perturbations observed across more training contexts are recovered significantly better than those seen in few contexts, while raw cell count correlates only weakly with recovery.

We highlight one structural point that, to our knowledge, none of these three papers can make from their own data: under the two-term parametric loss of \citet{hoffmann2022} in Eq.~\ref{M-eq:chinchilla}, increasing $D$ at \emph{fixed or uncontrolled} $N$ is \emph{expected} to plateau once the data term $B/D^\beta$ shrinks below the capacity term $A/N^\alpha$; the model lacks the capacity to exploit more data. That is a capacity bottleneck, rather than proof that a data-scaling law does not exist. Distinguishing the two requires a joint grid in which $N$ and $D$ are both swept independently, a design lacking in both the data sweep over fixed or uncontrolled model size by \citet{denadel2026} and the diversity sweep at a fixed model size in \citeauthor{wang2026xcell}'s appendix. While these studies conclude that data scaling is absent or saturates early, our work demonstrates that a joint scaling law holds and fits the empirical data well once model capacity and data budget are varied independently, distinguishing capacity limits from data limits.

\section{Experimental Design: A Controlled $N\times D$ Scaling Grid}
\label{sec:exp-design}

This appendix specifies the grid whose design and motivation are given in Section~\ref{M-sec:overview-grid}.

\subsection{Controlling for model shape}
A naive sweep over parameter count alone confounds size with shape (depth vs.\ width), since many $(d_{\text{model}}, n_{\text{layer}})$ pairs give the same parameter count. We instead fix the shape of the network across sizes: a constant width/depth ratio $d_{\text{model}}/n_{\text{layer}} = 128$, a constant head dimension of $64$, and an MLP expansion ratio fixed at $d_{\text{ffn}} = 2\,d_{\text{model}}$, so that moving along the $N$ axis scales width and depth together in fixed proportion. We verify the exact parameter count by instantiating each model rather than relying on a closed-form estimate. Table~\ref{tab:archs} gives the five configurations used on the $N$ axis of the grid; they span a factor of $\approx 64\times$ in parameter count.

\begin{table}[t]
\centering
\begin{tabular}{@{}lrrrrr@{}}
\toprule
Label & $d_{\text{model}}$ & $n_{\text{head}}$ & $n_{\text{layer}}$ & $d_{\text{ffn}}$ & Actual params \\
\midrule
1.31M & 256 & 4  & 2 & 512  & 1{,}311{,}744 \\
4.43M & 384 & 6  & 3 & 768  & 4{,}425{,}984 \\
10.5M & 512 & 8  & 4 & 1024 & 10{,}489{,}856 \\
20.5M & 640 & 10 & 5 & 1280 & 20{,}486{,}400 \\
83.9M & 1024 & 16 & 8 & 2048 & 83{,}902{,}464 \\
\bottomrule
\end{tabular}
\caption{Transformer configurations on the $N$ axis of the grid, at fixed shape (width/depth ratio $=128$, head dimension $=64$, $d_{\text{ffn}}=2\,d_{\text{model}}$).}
\label{tab:archs}
\end{table}

\subsection{Grid and per-point hyperparameter tuning}
The five model sizes of Table~\ref{tab:archs} are crossed with data budgets $D$ from $38.1$M to $362$M cells. Rather than a full dense crossing, we place points to give broad, independent variation in both axes (each model size is trained at several $D$, and each $D$ regime is covered by more than one model size), yielding $22$ distinct $(N,D)$ points for the fit, which is more than enough to identify the five parameters of Eq.~\ref{M-eq:chinchilla} while keeping total compute small. The experiments reported here were run on a single server with one NVIDIA H100 NVL GPU (approximately 96 GB of memory), 40 AMD EPYC CPUs, and 314 GiB of system RAM. Each run uses batch size $256$, dropout $0.1$, warmup ratio $0.1$, and cosine decay to a minimum learning rate of one-tenth of the run's peak rate (a $10\times$ decay, following~\cite{rae2021}); a run's step budget is set directly by its $D$ (Section~\ref{sec:data}). Because the optimal learning rate shifts with model size, we do not reuse a single global rate: for each model size we run a short learning-rate search (a handful of candidate rates) and keep the rate with the best loss, which we then reuse across all data sizes for that model.

\subsection{Parametric fit (Chinchilla Approach 3)}
\label{sec:fit}
For each $(N,D)$ point we record the self-supervised pretraining loss $L(N,D)$ and fit Eq.~\ref{M-eq:chinchilla} by nonlinear least squares on the log-residuals $\log \hat{L}(N,D;\theta) - \log L(N,D)$, with a Huber loss ($\delta=10^{-3}$) in place of a standard quadratic to limit the influence of outlier points. Because Eq.~\ref{M-eq:chinchilla} is non-convex in $\theta = (\log A, \log B, \log E, \alpha, \beta)$, we repeat the fit from a $5\times5\times5\times6\times6 = 4{,}500$-point grid of initializations ($\alpha_0,\beta_0 \in \{0,0.5,1,1.5,2\}$, $\log E_0 \in \{-1,-0.5,0,0.5,1\}$, $\log A_0,\log B_0 \in \{0,5,10,15,20,25\}$) and keep the lowest-cost solution. We report the fitted $(A,B,E,\alpha,\beta)$, the log-loss RMSE and $R^2$, and a residual plot (Appendix~\ref{app:residuals}) as fit diagnostics. As a stronger check than in-sample diagnostics alone, we additionally hold four $(N,D)$ points out of the fit entirely and compare their measured losses to what the fitted law predicts (Section~\ref{M-sec:results}).

\subsection{Compute-optimal frontier}
Approximating training compute as $C \approx 6ND$ (with $D$ counted in pretraining \emph{cells processed}, the training-token analogue of Section~\ref{sec:data}) and substituting $D = C/6N$ into Eq.~\ref{M-eq:chinchilla}, the loss-minimizing model size at fixed compute has the closed form
\begin{equation}
N^\ast(C) = \left(\frac{\alpha A}{\beta B}\right)^{\frac{1}{\alpha+\beta}} \left(\frac{C}{6}\right)^{\frac{\beta}{\alpha+\beta}},
\label{eq:frontier}
\end{equation}
equivalently the locus where the marginal returns to $N$ and $D$ balance ($\alpha A/N^{\alpha} = \beta B/D^{\beta}$), which we plot against iso-loss contours of the fitted surface (Figure~\ref{M-fig:isoloss}) as our primary scaling-law figure.

\section{Supplementary diagnostics}

\subsection{Fit residual diagnostics}
\label{app:residuals}

\begin{figure}[t]
\centering
\includegraphics[width=\columnwidth]{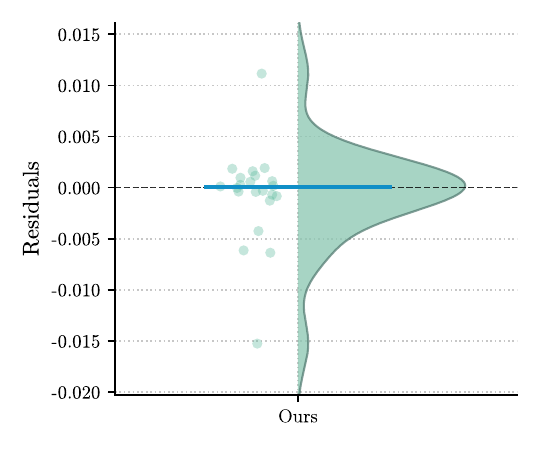}
\caption{Distribution of log-residuals $\log \hat{L} - \log L$ of the
Approach-3 fit over the $22$ grid points. Residuals are small ($\lesssim 2\%$) and roughly centered on zero, with no systematic structure.}
\label{fig:residuals}
\end{figure}

Figure~\ref{fig:residuals} shows the distribution of log-residuals $\log \hat{L} - \log L$ of the Approach-3 fit (Section~\ref{sec:fit}) over the $22$ grid points. The residuals are small ($\lesssim 2\%$), roughly centered on zero, and show no systematic structure, representing the in-sample diagnostic supporting the fit quality reported in Section~\ref{M-sec:results}.

\subsection{Proportional vs.\ uniform pool sampling}
\label{app:sampling}

\begin{figure}[t]
\centering
\includegraphics[width=\columnwidth]{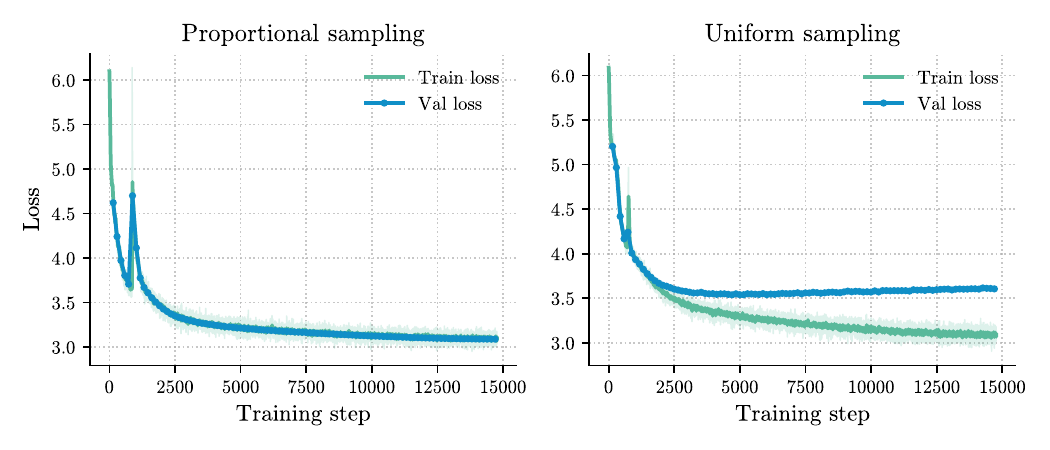}
\caption{Training and validation loss for two pretraining runs differing in how the cell-type pool behind each sentence is sampled. \textbf{Left:} proportional to the pool's cell count, where validation tracks training for the whole run and is still at its minimum at the last step. \textbf{Right:} uniform over cell types, where training keeps falling, but validation flattens at $\approx\!3.54$ around step $4{,}700$ and then rises, the signature of memorizing the oversampled rare types. The runs differ only in the pool weighting. Training loss is shown raw (light curves) and EWMA-smoothed (solid). Absolute losses are not comparable across panels, as each run trains and validates under its own sampling scheme, so the overfitting evidence is the within-run divergence; see the caveat above.}
\label{fig:sampling}
\end{figure}

Figure~\ref{fig:sampling} contrasts two pretraining runs that differ in how the cell-type pool behind each sentence is chosen (Section~\ref{sec:data}), and motivates our use of proportional sampling throughout.

Under proportional sampling (left), where a pool is drawn with probability proportional to its cell count, the validation loss tracks the training loss for the whole run: the two end at $3.09$ and $3.10$ respectively, a negligible generalization gap, and the validation loss is still at its minimum at the last measured step; the run is data-limited, not overfitting. Under uniform sampling (right), where every cell type is equally likely regardless of how many cells it holds, the curves separate after $\approx\!2{,}500$ steps. The training loss continues down to $\approx\!3.09$, showing that uniform sampling is not failing to \emph{optimize}, fitting its own training distribution about as far as the proportional run fits its own, while the validation loss flattens near $3.54$ at $\approx\!4{,}700$ steps and then \emph{rises} to $3.61$ by the end of the budget. A training loss that keeps falling while the validation loss turns upward is the standard signature of overfitting, and it is what uniform weighting should produce on a corpus whose cell types are long-tailed: equal weight per type oversamples the rare types, so the few cells they hold recur many times within a run and are memorized rather than generalized from.

The two runs are a controlled ablation of the pool weighting: they share the same corpus, model, step budget, and seed, and differ only in whether the cell-type pool is drawn proportionally to its cell count or uniformly.

One caveat on how the figure should be read. Absolute losses are not comparable \emph{across} the two panels, because the pool weighting is a property of the sentence sampler and that sampler serves both splits: each run draws its validation sentences under the same weighting as its training sentences, so the uniform run is scored on a rebalanced, intrinsically harder validation distribution. Each panel therefore reports a loss on its own run's distribution, and the evidence for overfitting is the \emph{within-run} divergence between the two curves, where training and validation do share a distribution, rather than the gap between the panels.

\subsection{Held-out perturbation-benchmark loss during pretraining}
\label{app:checkmark}

Alongside the scBaseCount validation split the fit uses, every pretraining run also scores next-token loss on a held-out pool of control (``non-targeting'') cells drawn from the perturbation-response benchmark a Stage 2 finetune would target (Section~\ref{M-sec:discussion}), using the same unconditioned, same-cell-type sentence format as pretraining --- the perturbation identity is never shown to the model at this stage. This gives a second, narrower generalization probe alongside the primary scBaseCount loss, tracked throughout every run, and is the basis for the observation discussed in Section~\ref{M-sec:discussion}.

\begin{figure}[t]
\centering
\includegraphics[width=\columnwidth]{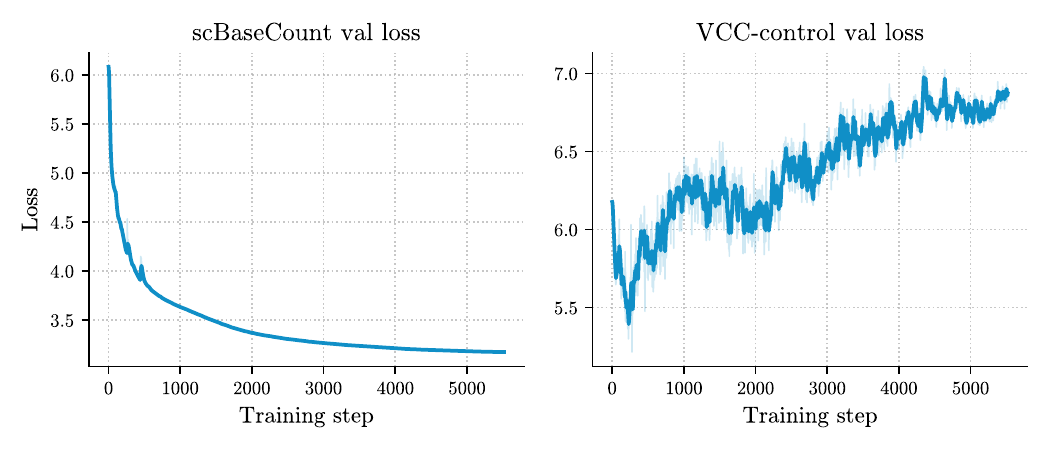}
\caption{Two losses from the same pretraining run, representative of the pattern seen at every model and data size in the grid. \textbf{Left:} the scBaseCount validation loss the scaling law is fit to, which decreases smoothly throughout training. \textbf{Right:} held-out next-token loss on control cells from the perturbation-response benchmark, which instead falls sharply early in training and then rises for the remainder of the run --- a checkmark shape. Raw per-step values shown faint, EWMA-smoothed on top.}
\label{fig:checkmark}
\end{figure}

Figure~\ref{fig:checkmark} plots both losses side by side over one representative pretraining run; we observe the same pattern at every model and data size in the grid (Section~\ref{sec:exp-design}). The scBaseCount validation loss (left) decreases smoothly throughout training and is still falling at the end of the run, consistent with the proportional-sampling behavior of Appendix~\ref{app:sampling}. The held-out perturbation-benchmark loss (right), scored on the same run at the same steps, instead falls sharply early in training, reaches a minimum well before the run ends, and then rises for the remainder of training, finishing above where it started. The two losses are on the same scale (next-token cross-entropy under the same tokenizer) but diverge in direction: continued pretraining keeps improving the fit to the broad scBaseCount distribution while making the fit to this narrower, perturbation-benchmark-specific distribution worse.

\subsection{Per-checkpoint generation quality}
\label{app:gen-trend}

Table~\ref{tab:gen-trend} gives the concrete per-checkpoint values plotted in Figure~\ref{M-fig:gen-trends}: generation quality for the $4.43$M model as its pretraining validation loss falls over training.

\begin{table}[t]
\centering
\small
\begin{tabular}{@{}rrccc@{}}
\toprule
Step & Val.\ loss & Pearson & MAE & Disc. \\
\midrule
1{,}470  & 3.610 & 0.9467 & 0.0515 & 0.9419 \\
2{,}940  & 3.309 & 0.9549 & 0.0466 & 0.9679 \\
4{,}410  & 3.231 & 0.9554 & 0.0455 & 0.9752 \\
5{,}880  & 3.193 & 0.9575 & 0.0441 & 0.9804 \\
7{,}350  & 3.167 & 0.9584 & 0.0437 & 0.9761 \\
8{,}820  & 3.142 & 0.9595 & 0.0422 & 0.9867 \\
10{,}290 & 3.122 & 0.9597 & 0.0423 & 0.9832 \\
11{,}760 & 3.106 & 0.9599 & 0.0415 & 0.9869 \\
13{,}230 & 3.095 & 0.9604 & 0.0412 & 0.9914 \\
14{,}700 & 3.090 & 0.9605 & 0.0412 & 0.9896 \\
\midrule
\multicolumn{2}{@{}l}{Baseline}  & 0.9637 & 0.0382 & 0.9978 \\
\bottomrule
\end{tabular}
\caption{Per-checkpoint generation quality for the $4.43$M model as its pretraining validation loss falls over training, representing the concrete values plotted in Figure~\ref{M-fig:gen-trends}. Pearson, MAE, and Disc.\ are the mean-expression Pearson correlation, the mean-expression mean absolute error, and the cell-type discriminability score. Lower loss yields higher Pearson and discriminability and lower MAE. The baseline is repeated from Table~\ref{M-tab:generation} as a reference bound.}
\label{tab:gen-trend}
\end{table}

\section{Limitations}
\label{sec:limitations}

Our conclusions are bounded by the extent of the grid and by the objective we fit. The parameter axis spans $1.3$--$83.9$M, well below the largest models in Table~\ref{M-tab:related}, so we cannot rule out a change in the exponents at much larger scale; the data axis, from $38.1$M to $362$M vectors, is comparable to or larger than the corpora used in several of those studies. The law is fit to the self-supervised pretraining loss on held-out context sequences, so we make no claim about how downstream, task-specific losses scale, and it is fit for a single architecture family, a causal transformer over a discrete, learned RQ-VAE tokenization, so transfer to continuous-embedding models is untested. Our compute proxy $C\approx6ND$ counts $D$ in expression vectors rather than tokens, which offsets our FLOP counts from token-based ones by a roughly constant factor. Finally, the tokenizer is lossy by construction, placing an upper bound on reconstruction fidelity independent of transformer capacity, and Approach 3 fits are sensitive to the initialization grid, which we mitigate with a wide multi-start grid, detailed in Section~\ref{sec:fit}, and with the out-of-sample held-out points of Section~\ref{M-sec:heldout}.

\end{document}